\documentclass{article}

\PassOptionsToPackage{numbers, compress}{natbib}

\usepackage[preprint]{neurips_2019} 

\usepackage[utf8]{inputenc} 
\usepackage[T1]{fontenc}    
\usepackage{hyperref}       
\usepackage{url}            
\usepackage{booktabs}       
\usepackage{amsfonts}       
\usepackage{nicefrac}       
\usepackage{microtype}      

\usepackage{afterpage}

\usepackage{amssymb}
\usepackage{amsmath}
\usepackage{enumitem}
\usepackage{booktabs}
\usepackage{multirow}

\usepackage{lipsum}

\usepackage{colortbl}

\DeclareMathOperator{\EX}{\mathbb{E}}

\usepackage[small]{caption}
\usepackage{graphicx}
\usepackage[table]{xcolor}

\usepackage{appendix}

\newcommand\R{{\mathbb R}}

\DeclareMathOperator*\argmin{{arg\,min}}

\newcommand{\nosemic}{\renewcommand{\@endalgocfline}{\relax}}
\newcommand{\dosemic}{\renewcommand{\@endalgocfline}{\algocf@endline}}
\let\oldnl\nl
\newcommand{\nonl}{\renewcommand{\nl}{\let\nl\oldnl}}

\usepackage{array}
\usepackage{float}
\usepackage[linesnumbered,ruled]{algorithm2e}
\usepackage{setspace}

\usepackage{enumitem}

\newtheorem{theorem}{Theorem}

\title{Learning Surrogate Losses}

\author{%
Josif Grabocka, Randolf Scholz, Lars Schmidt-Thieme \\
  ISMLL, University of Hildesheim\\
  31141 Hildesheim, Germany\\
  \texttt{\{josif, scholz, schmidt-thieme\}@ismll.uni-hildesheim.de} 
}

\begin{document}

\maketitle

\begin{abstract}

The minimization of loss functions is the heart and soul of Machine Learning. In this paper, we propose an off-the-shelf optimization approach that can minimize virtually any non-differentiable and non-decomposable loss function (e.g. Miss-classification Rate, AUC, F1, Jaccard Index, Mathew Correlation Coefficient, etc.) seamlessly. Our strategy learns smooth relaxation versions of the true losses by approximating them through a surrogate neural network. The proposed loss networks are set-wise models which are invariant to the order of mini-batch instances. Ultimately, the surrogate losses are learned jointly with the prediction model via bilevel optimization. Empirical results on multiple datasets with diverse real-life loss functions compared with state-of-the-art baselines demonstrate the efficiency of learning surrogate losses.

\end{abstract}

\section{Introduction}

\begin{figure}[t] 
	\centering
	\includegraphics[width=0.9\linewidth, trim={2.5cm 5.2cm 2.6cm 1.0cm}]{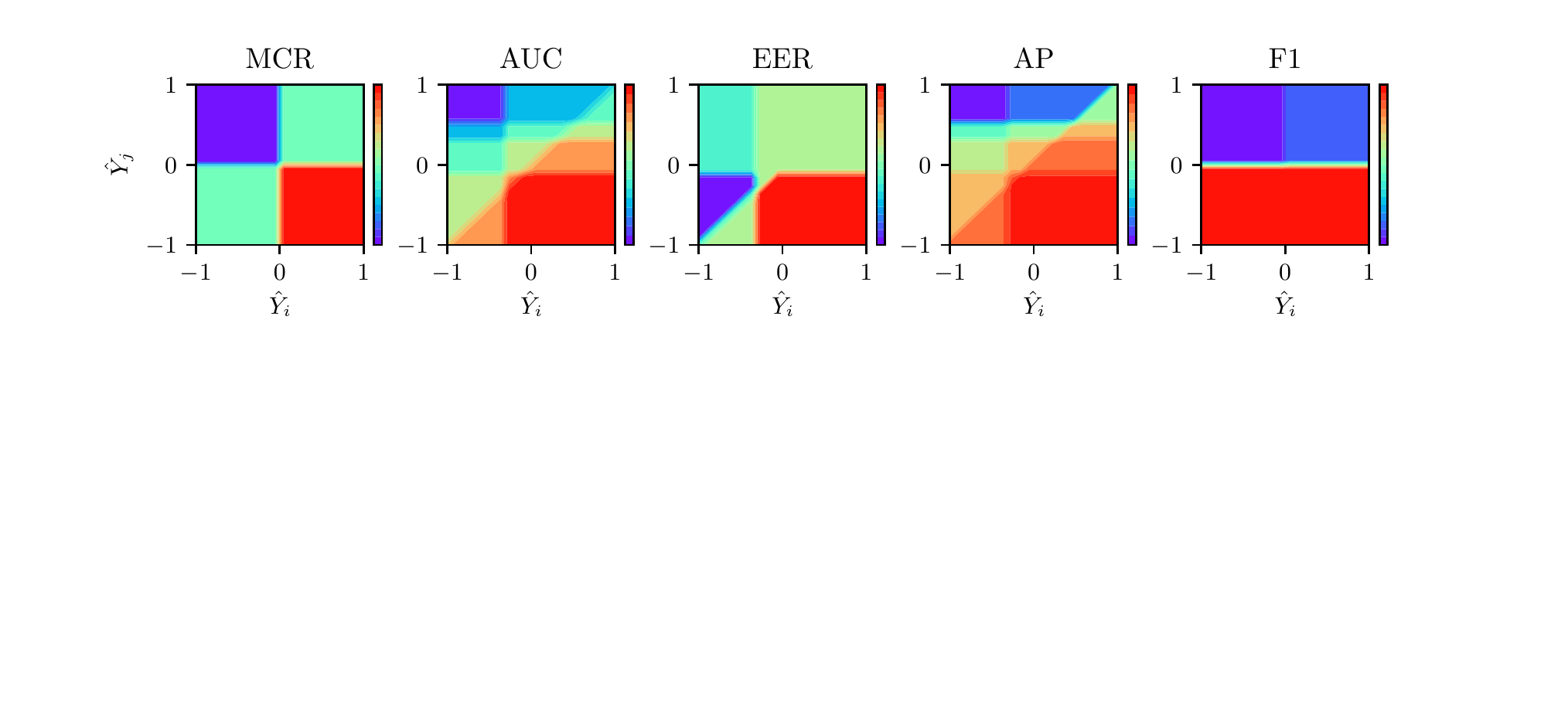}
	\caption{The surfaces of five binary classification losses derived by perturbing the predictions $\hat Y_i, \hat Y_j$ of two instances inside a random mini-batch $\hat Y \in \R^{10}, Y \in \{0,1\}^{10}$, with true targets $Y_i=0, Y_j=1$. AUC, AP and F1 are converted to losses via $1-x$, while the positive class for MCR, EER and F1 is estimated as $\hat Y \ge 0$. The illustration aims at showing the highly non-differentiable and piece-wise continuous loss surfaces.}
	\label{fig:hardlosses}
\end{figure}

In reality, a large set of loss functions cannot be directly minimized by gradient-based methods because they are either piece-wise continuous, non-differentiable, or non-decomposable~\cite{DBLP:conf/nips/ZhangLZY18}. For example, binary classification models are often evaluated using the miss-classification rate (MCR), Area under the ROC curve (AUC), F1 measure (F1), Jaccard Similarity Index (JAC), Average Precision (AP), Equal Error Rate (EER), or the Mathew Correlation Coefficient (MCC). On the other hand, learning-to-rank models are measured through the Normalized Discounted Cumulative Gain (NDCG), or the Mean Average Precision (MAP). To illustrate the point, Figure~\ref{fig:hardlosses} shows the challenging surfaces of common binary classification losses. Unfortunately, there exists no tractable omni-solver so far, i.e. no off-the-shelf optimization strategy to train prediction models for the aforementioned category of loss functions. It is worth pointing out that non-gradient-based approaches, such as evolutionary computing, are intractable from a runtime perspective.

While there exists a plethora of methods that tackle various aspects of particular losses, there is no single gradient-based method that can optimize any loss in a seamless manner. Researchers have so far focused on deriving smooth surrogate relaxations (approximations) to the true loss functions~\cite{berman2018lovasz,pmlr-v54-eban17a}, in a way that first- and second-order optimization techniques can benefit from the derivative of the surrogates with respect to the parameters of prediction models. For instance, the widely-applied cross-entropy loss is a surrogate relaxation of the miss-classification rate. However, these explicit relaxations are hand-crafted individually for each loss and do not generalize to other losses.

In this paper, we present the first off-the-shelf optimizer for arbitrary loss functions. In contrast to the related work, this work proposes a new perspective on minimizing loss functions, by defining surrogate losses as meta-level neural networks that approximate the desired true non-differentiable losses. Our method does not need the gradient information of the true loss with respect to the parameters of the prediction model and treats the loss as a black-box function. In addition, we introduce a set-based surrogate network that computes the loss over the training set, being invariant to the order of instances, in order to accurately handle non-decomposable losses.

The surrogate learning problem is formalized as a bilevel programming task~\cite{colson2007overview,pmlr-v80-franceschi18a} that is trained through a concurrent optimization algorithm. This paper shows that universal surrogates, which are trained without paying attention to a specific dataset, are sub-optimal compared to surrogates learned in a per-dataset manner. Results on nine datasets demonstrate that learning surrogates produces more accurate prediction models than state-of-the-art baselines with regard to diverse loss functions.

\section{Related Work}

Due to the non-differential and non-continuous nature of most real-life losses, early works deployed the proxies of the miss-classification rate (e.g. cross-entropy) as universal proxy losses, despite their sub-optimal performance~\cite{cortes2004auc}. Subsequent approaches relied on designing smooth relaxations of the losses. As an example, the pairwise ranking loss is a common surrogate of the AUC measure~\cite{gao2015consistency,NIPS2009_3708}. On the other, hand, the F-measure is another typical loss that cannot be optimized directly due to its non-decomposable nature over instances. The initial papers tackling F1 focused instead on the empirical utility maximization paradigm~\cite{Ye:2012:OFT:3042573.3042772}. Later on, researchers addressed F1 by optimizing the hyper-parameters of standard binary classifiers, either the cost-sensitive weights of the classification loss~\cite{NIPS2014_5508}, or the threshold of the estimated target values~ \cite{10.1007/978-3-662-44851-9_15,NIPS2014_5454}. Nevertheless, the non-decomposable trait of F1 remains unresolved~\cite{DBLP:conf/nips/ZhangLZY18} and recent works have explored directions to improve hyper-parameter tuning with tighter bounds~\cite{pmlr-v89-bascol19a}. 

Instead of relying on explicit surrogates, another research direction handles non-convex losses by means of the \textit{direct loss} method~\cite{NIPS2010_4069}, which minimizes a surrogate loss by embedding the true loss as a correction term. This method was recently extended to optimize neural networks~\cite{pmlr-v48-songb16}. It assumes the loss can be decomposed into per-instance sub-losses and the authors derived an explicit decomposition of the average precision~\cite{pmlr-v48-songb16}. Unfortunately, per-instance dis-aggregations are not trivially feasible in other cases (e.g. F1), making the direct loss optimization technique an impractical off-the-shelf option.

Two more recent papers have offered relaxation surrogates for non-decomposable losses, concretely the AUC and the Jaccard Index (known as Intersection over Union in the computer vision community). The first method defines relaxation forms for the building blocks of a confusion matrix (e.g. true positives, true negatives etc.) and combines the building block relaxations to create a final surrogate for losses like the AUC~\cite{pmlr-v54-eban17a}. However, this model does not handle cases where the loss is not expressible into confusion matrix blocks, for instance the Jaccard index. On the other hand, the second paper proposes the Lovasz soft-max as a smooth approximation to the Jaccard index~\cite{berman2018lovasz}, which is based on a generic decomposition of sub-modular (decomposable) losses for sets~\cite{pmlr-v37-yub15}.

Besides that, a stream of recent papers has focused on meta-learning for loss functions. The "Learning to Teach" paradigm~\cite{DBLP:journals/corr/abs-1805-03643} proposes a meta-level teacher/controller that continuously updates the loss function for a prediction model based on its progress. The work has been recently extended to enable a gradient-based learning of the teacher/controller~\cite{learntoteachdynamic}. However, this approach does not extend to non-decomposable loss functions which are defined over the full set of instances. A parallel stream of research has elaborated the concept of "Learning to Learn"~\cite{DBLP:journals/corr/LiM16b}, or "Learning to Optimize"~\cite{pmlr-v70-chen17e}, which proposes to directly learn the amount of update values that are applied to the parameters of the prediction model. In the proposed framework a controller uses per-parameter learning curves comprised of the loss values and derivatives of the loss with respect to each parameters~\cite{pmlr-v70-chen17e}. This method suffers from two drawbacks that prohibit its direct applicability to arbitrary losses: i) for large prediction models it is computationally infeasible to store the learning curve of every parameter, and ii) there is no gradient information for non-differentiable losses.

In contrast to the prior work, we propose the first off-the-shelf optimization method that seamlessly minimizes any loss function. In Section~\ref{sec:sotacomp} we empirically compare the proposed method against multiple state-of-the-art relaxations, with regards to minimizing popular binary classification losses, such as AUC, F1, Jaccard Index and the miss-classification rate.

\section{Surrogate Learning}

Data mini-batches $(x,y) := \{(x_1,y_1), \dots, (x_N, y_N)\}$ of $N$ instances each, with features $x \in \R^{N \times M}$ and ground-truth targets $y \in \mathcal{Y}^{N}$, are drawn from a dataset $\mathcal{D}$ with a sampling distribution $P_\mathcal{D}\left(x,y\right)$, where typically $P$ is the uniform distribution. The target domain can be binary $y \in \{0,1\}^N$, or nominal among $C$ categories $y \in \{1,\dots,C\}^N$. Ultimately, the purpose of a prediction model is to estimate a target variable $\hat y(x; \alpha)\colon \R^{N\times M} \rightarrow \R^N$, where the prediction model has parameters $\alpha$. The estimations $\hat y \in \R^N$ need to accurately match the given ground-truth target variable $y \in \mathcal{Y}^{N}$ with regards to a desired loss function $\ell\left(y, \hat y\right)\colon \R^{N} \times \mathcal{Y}^{N} \rightarrow \R$. Therefore, supervised learning focuses on computing the optimal parameters $\alpha^{*}$ that minimize the following objective.

\begin{eqnarray}
\label{eq:supervisedlearning}
\alpha^{*} = \argmin_{\alpha} & \EX\limits_{(x,y) \sim P_{\mathcal{D}}\left(x,y\right) } &  \; \ell \left(y, \hat y(x; \alpha) \right) 
\end{eqnarray}

To minimize the aforementioned objective through first- or second-order optimization, it is necessary to define the gradients $\frac{\partial \ell}{\partial \alpha}$. Unfortunately, in most real-world cases the loss represents step functions that are only piece-wise continuous (MCR, F1, AUC, etc.). Therefore, the derivatives are either zero, or undefined at the function steps, which prohibits a direct optimization of these losses. For this reason, a smooth surrogate relaxations of the loss functions is used instead of the true loss. Arguably the most popular surrogate is cross-entropy, which is a relaxation of the miss-classification rate. Yet, such non-parametric relaxation functions are not trivially derivable when the loss is non-decomposable into per-instance components (e.g. F1, AUC), because such losses are defined as performance measures over an entire set of instances. 

Instead of deriving one explicit hand-crafted function $\hat \ell$ for the surrogate of every loss function $\ell$, we propose a method that \textit{parameterizes} and \textit{learns} the surrogate for any demanded loss function through an off-the-shelve procedure. Neural networks are a good choice for parametrizing the surrogate loss $\hat \ell$ given their universal approximation capability~\cite{HORNIK1991251}. However, the surrogate loss network must be a permutation-invariant set model, whose output must remain fixed given different orders of instances within a dataset mini-batch, i.e. $\hat \ell\left(y_1, \dots, y_N, \hat y_1, \dots, \hat y_N \right) = \hat \ell\left(y_{\pi\left(1\right)}, \dots, y_{\pi\left(N\right)}, \hat y_{\pi\left(1\right)}, \dots, \hat y_{\pi\left(N\right)}\right)$ for any index permutation $\pi$. Our method uses the Kolmogorov–Arnold representation theorem~\cite{tikhomirov1991representation} and defines the surrogate loss of Equation~\ref{eq:pointwiseModel} as a composition function $h$ of per-instance functions $g$. This type of aggregation was recently applied to learning neural networks over sets of instances~\cite{NIPS2017_6931}.

\begin{eqnarray}
\label{eq:pointwiseModel}
\hat \ell\left(y, \hat y \right) = h\left( \frac{1}{N} \sum_{i=1}^{N} g \left(y_i, \hat y_i \right)  \right)
\end{eqnarray}

where $g\colon \R^2 \rightarrow \R^{Q}$, and $h\colon \R^{Q} \rightarrow \R$ are deep forward neural networks. The first function $g$ extracts $Q$ latent error components for the predictions on each instance (e.g. latent representations of false positive, false negative, etc.), while the aggregation produces set-wise performance indicators (e.g. potentially latent representation of the count of true positives, false positives, error rate, etc.). The function $h$ creates nonlinear combinations of set-wise performance indicators to produce complex latent error metrics such as precision, recall, F1, etc. It has been proven that with sufficient capacities for $g$ and $h$, any permutation-invariant set function (hence any loss) can be approximated this way~\cite{NIPS2017_6931}. The remaining sections of this paper detail a novel method for optimizing surrogate losses. 

\subsection{Universal Surrogates}
\label{sec:universalsurrogate}

The first intuition is to learn the weights $\beta$ of surrogate network (a.k.a. the functions $g$ and $h$) in a universal manner by solving the objective of Equation~\ref{eq:LUniversal}. In other words, we can attempt to make any surrogate loss $\hat \ell$ behave as any true loss $\ell$ over the space of all possible batches of randomly-drawn true $y$ and estimated $\hat y$ targets. Focusing on binary classification losses, we sample the ground truth from a Bernoulli distribution, and respective estimated targets from a Gaussian distribution. Furthermore, the meta-loss $\mathcal{L}: \R \times \R \rightarrow \R$ measures the distance between the true and surrogate losses and is practically implemented as an $L_1$ norm.

\begin{eqnarray}
\label{eq:LUniversal}
\beta^{\text{Universal}} =\argmin_{\beta} & \EX\limits_{\, y \sim \mathcal{B}\left(p\right)_{N \times 1}, \; \hat y \sim \mathcal{N}\left(\mu, \sigma^2\right)_{N \times 1} }  & \mathcal{L}\left( \ell\left(y, \hat y\right), \hat \ell\left(y, \hat y  \, ; \, \beta \right)  \right)
\end{eqnarray}

The challenge of learning a universal surrogate is in ensuring that the sampling hyper-parameters ($p, \mu, \sigma$) would lead to drawing $(y,\hat y)$ that match the specific target distribution of a concrete dataset. Therefore, we empirically found out that it is sub-optimal to universally relax the whole space of the true loss. Instead, as future sections will detail, it is more efficient to design a relaxation in a dataset-specific manner, which smoothens only the specific regions of the true loss which belong to the respective dataset-specific distributions of true and estimated target batches. 

\subsection{Surrogate Learning as Bilevel Programming}
\label{sec:bileveloptim}

A different approach from the universal surrogate is to associate a surrogate loss to each dataset. In that manner, the surrogate creates a smooth relaxation only around the dataset-relevant regions of the target $(y, \hat y(x; \alpha))$ space. We propose to jointly optimize the prediction model parameters ($\alpha$) and the surrogate loss parameters ($\beta$) through the following optimization:

\begin{eqnarray}
\label{eq:LHat}
\alpha^{*} = \argmin_{\alpha} & \EX\limits_{(x,y) \sim P_{\mathcal{D}}\left(x,y\right) } & \hat \ell \left(y, \hat y \left(x \, ; \, \alpha \right); \, \beta \right)  \\ 
\label{eq:LApprox}
\beta^{*} =\argmin_{\beta} & \EX\limits_{(x,y) \sim P_\mathcal{D}\left(x,y\right) }  & \mathcal{L}\left( \ell \left(y, \hat y \left(x ; \alpha \right) \right), \hat \ell\left(y, \hat y \left(x ; \alpha \right); \, \beta \right) \right) 
\end{eqnarray}

The rationale is that Equation~\ref{eq:LHat} optimizes the prediction model in order to minimize the surrogate loss. However, the surrogate loss should approximate the true loss, which is ensured by Equation~\ref{eq:LApprox}. Considered in isolation, the two objectives have an interdependent and rivaling \textit{modus operandi}:

\begin{enumerate}[label=(\alph*)]
\item Equation~\ref{eq:LHat}: The model $\hat y$ is accurate only if $\hat \ell$ closely approximates $\ell$'s minima;
\item Equation~\ref{eq:LApprox}: The surrogate $\hat \ell$ approximates $\ell$'s local minimum only around the region yield by the given $\hat y$, because $\hat \ell$ depends on $\alpha$.
\end{enumerate}

\begin{figure}[t] 
	\centering
	\includegraphics[width=0.9\linewidth, trim={1.5cm 6.0cm 1.4cm 0.0cm}]{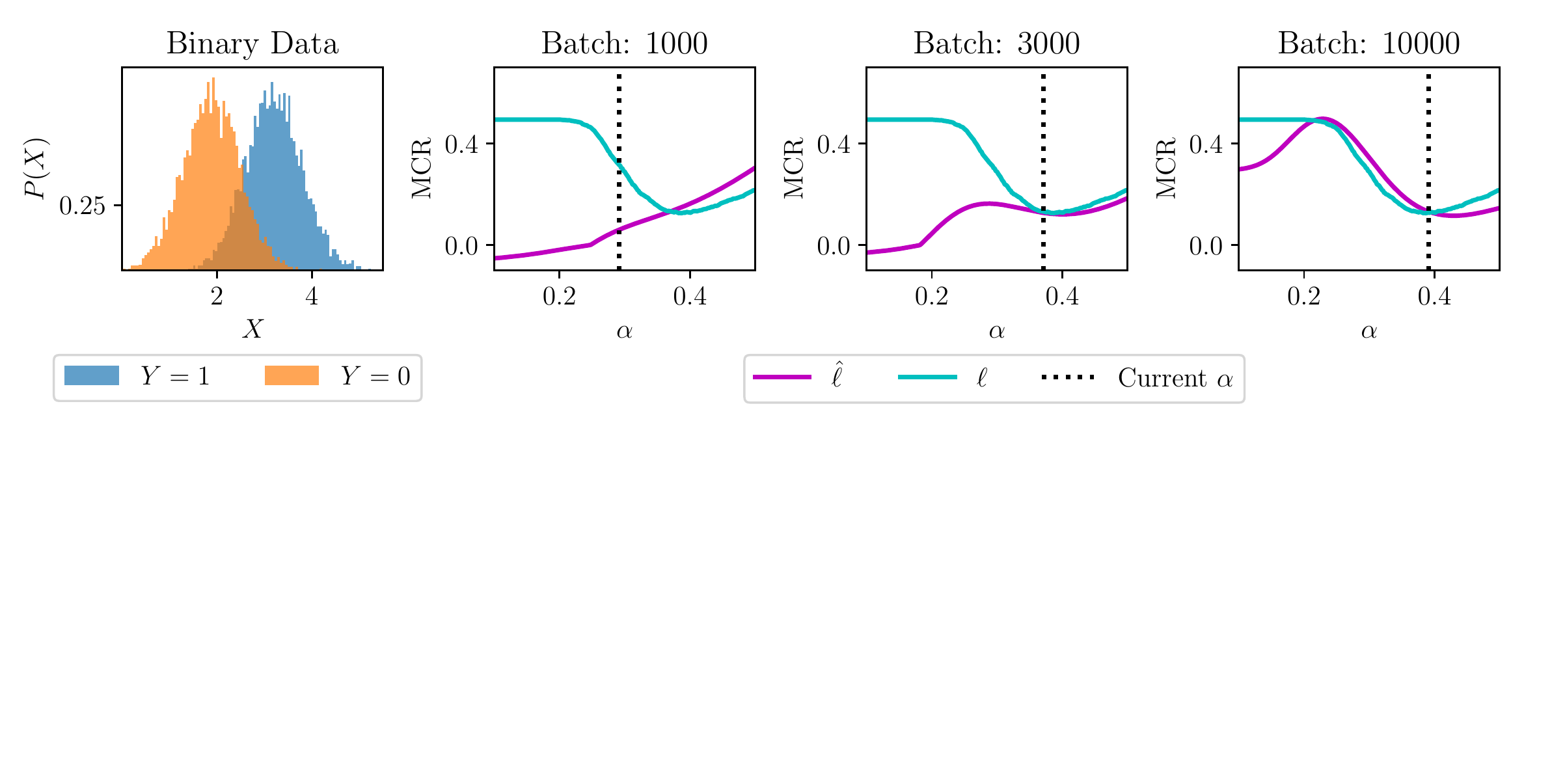}
	\caption{A single-parameter prediction model $\hat y(x, \alpha) = \alpha x - 1$ classifies a single-feature binary dataset (leftmost plot), i.e. $\alpha \in \R, x \in \R^{N \times 1}$, where initially $\alpha=0.3$. In the three rightmost plots, the x-axis shows the variation in true (cyan) and surrogate (magenta) losses for the whole space of $\alpha$. The plots indicate that Equation~\ref{eq:LApprox} forces the surrogate to approximate the true loss at the regions around the current $\alpha$ (dashed vertical line), while the parameter $\alpha$ is updated towards the minimum of the surrogate by Equation~\ref{eq:LHat}.}
	\label{fig:minimalistic}
\end{figure}

However, the interdependence can be addressed by treating both objectives of Equations~\ref{eq:LHat} and \ref{eq:LApprox} as a concurrent relationship, in a way that we can optimize them jointly and simultaneously. This dual objective is an instance of a bilevel programming problem~\cite{colson2007overview,pmlr-v80-franceschi18a}. Algorithm~\ref{alg:learning} sketches the minimization procedure of the proposed surrogate learning. In an alternating fashion, the surrogate minimization is carried out for $K_\alpha$ steps (lines $4-7$), while the surrogate loss is updated for $K_\beta$ steps (lines $9-12$).  To illustrate the mechanism, Figure~\ref{fig:minimalistic} provides a minimalistic example of optimizing jointly the bilevel programming objectives for training a single-parameter model on a single-feature binary dataset. The plots show the effect of minimizing the surrogate (magenta), as the parameter $\alpha$ is updated towards the minimum of the surrogate. At the same time, we observe that the surrogate is updated to match the true miss-classification-rate loss (cyan) at the current parameter value. The model converges close to the optimal true loss after applying the steps of Algorithm~\ref{alg:learning}. 

\begin{algorithm}
	\SetAlgorithmName{Algorithm}{}{}
	\SetKwInOut{Return}{Return}
	\SetKwInOut{Input}{Input}
	\Input{$\;\;$Dataset $\mathcal{D}$, Loss $\ell$, Training epochs $T$, Update steps $K_\alpha, K_\beta$, Learning rates $\eta_{\alpha}, \eta_{\beta}$.}
	Initialize $\alpha$, $\beta$ \\
		\For{$1, \ldots, T$}
		{
			$\alpha^{(0)} \leftarrow \alpha $ \\
			\For{$k=1, \ldots, K_{\alpha}$} 
			{
				Sample batch $(x,y) \sim P_\mathcal{D}\left(x,y\right)$ \\
				$\alpha^{(k)} \leftarrow \alpha^{(k-1)} - \eta_{\alpha^{(k-1)}} \nabla_{\alpha} \hat \ell \left(y, \hat y \left(x \, ; \, \alpha^{(k-1)} \right); \, \beta \right) $
			}
			$\alpha \leftarrow \alpha^{(K_{\alpha})}, \beta^{(0)} \leftarrow \beta$ \\
			\For{$k=1, \ldots, K_{\beta}$}
			{
				Sample batch $(x,y) \sim P_\mathcal{D}\left(x,y\right)$ \\
				$\beta^{(k)} \leftarrow \beta^{(k-1)} - \eta_{\beta^{(k-1)}} \nabla_{\beta} \mathcal{L}\left( \ell \left(y, \hat y \left(x ; \alpha \right) \right), \hat \ell\left(y, \hat y \left(x ; \alpha \right); \, \beta^{(k-1)} \right) \right)  $
			}
			$\beta \leftarrow \beta^{(K_{\beta})}$
		 
		} 
	\caption{Surrogate Learning}
	\Return{$\;\;$Prediction model $\hat y\left(x; \, \alpha\right)$}
	\label{alg:learning}
\end{algorithm}

%
%


\subsection{Note on convergence}

The optimization of Equations~\ref{eq:LHat}-\ref{eq:LApprox} can be rewritten as a standard bilevel programming in Equation~\ref{eq:bilevelform}:   

\begin{eqnarray}
\label{eq:bilevelform}
\min_{\alpha \in \mathcal{A}} \; \hat \ell(\alpha, \beta^{*}\left(\alpha\right)), \;\;\;\; \beta^{*}\left(\alpha\right) = \argmin_{\beta\left(\alpha\right)} \; \mathcal{L}\left( \alpha, \beta\left(\alpha\right) \right)
\end{eqnarray}

Consequently, it is possible to show the convergence of our problem in the lines of existing converge proofs for bilevel programming~\cite{Revalski1993,pmlr-v80-franceschi18a}. Given the equivalence of our task and bilevel programming, the convergence of Algorithm~\ref{alg:learning} is stated in Theorem~\ref{theorem:conv}.

\begin{theorem}[Convergence]
\label{theorem:conv}
Considering the following assumptions~\cite{Revalski1993,pmlr-v80-franceschi18a}:
\begin{enumerate}[label=(\alph*)]
	\item The parameter space $\alpha \in \mathcal{A}$ is a compact subset and $\hat \ell\left(\alpha, \beta\left(\alpha\right)\right)$ is jointly continuous, while the mapping $\left(\alpha, \beta\left(\alpha\right)\right) \rightarrowtail \mathcal{L}\left( \alpha, \beta\left(\alpha\right) \right)$ is also jointly continuous;  
	\item $\argmin_{\beta\left(\alpha\right)} \mathcal{L}\left(\alpha, \beta\left(\alpha\right) \right)$ is a singleton for every $\alpha \in \mathcal{A}$, and \\ $\beta^{*}\left(\alpha\right) = \argmin_{\beta\left(\alpha\right)} \mathcal{L}\left( \alpha, \beta\left(\alpha\right) \right)$ is bounded as $\alpha$ varies in $\mathcal{A}$;
	\item $\mathcal{L}\left(\alpha, \beta\right)$ is Liptschitz continuous and $\beta^{(K_{\beta})}\left(\alpha\right) \to \beta^{*}\left(\alpha\right)$  uniformly for all $\alpha$ as $K_{\beta} \to \infty$.
\end{enumerate}
Then for $K_{\beta} \to \infty$ the following holds~\cite{pmlr-v80-franceschi18a}:
\begin{enumerate}[label=(\alph*)]
\item $\inf_{\alpha} \hat \ell(\alpha, \beta^{(K_{\beta})}\left(\alpha\right)) \rightarrow \inf_{\alpha} \hat \ell(\alpha, \beta^{*}\left(\alpha\right))$;  \; 
\item $\argmin_{\alpha} \hat \ell(\alpha, \beta^{(K_{\beta})}\left(\alpha\right)) \rightarrow \argmin_{\alpha} \hat \ell(\alpha, \beta^{*}\left(\alpha\right))$ in a notion of set convergence.
\end{enumerate}
\end{theorem}
\textbf{Proof:} Detailed in the related work~\cite{pmlr-v80-franceschi18a}.

\section{Experiments}

\subsection{Protocol}
\label{sec:protocol}

The experiments are focused on a collection of publicly-available binary classification datasets, whose statistics are presented in Table~\ref{tab:datasets}. We split the data randomly into $80\%$ training and $20\%$ testing instances. The prediction model $\hat y$ has an architecture of $[100, 30, 10, 1]$ neurons with Leaky-ReLU activation, where batch normalization was applied at each layer. In addition, after each batch normalization we added a drop-out regularization layer with the drop rate set to 0.2. The surrogate network component $g$ has $[30,30]$ neurons, while $h$ has $[10, 10, 1]$ neurons. We employed the Adam optimizer for training the network, with initial learning rates being $\eta_\alpha=\eta_\beta=10^{-5}$.

\begin{table}[h!]
	\centering 
	\small
	\caption{Dataset Statistics. }
	\label{tab:datasets}
	\begin{tabular}{ccrrrcc}
		\toprule \textbf{Dataset} & \textbf{Classes} & \textbf{Train} & \textbf{Test} & \textbf{Features} & \textbf{Pos. Frac.} \\ \midrule
		\text{A9A} & 2 & 39073 & 9769 & 123 & 0.2379 \\
		\text{CC-Fraud} & 2 & 227845 & 56962 & 29 & 0.0015 \\
		\text{Cod-RNA} & 2 & 390852 & 97713 & 8 & 0.3346\\
		\text{CoverType} & 2 & 464809 & 116203 & 54 & 0.4867 \\
		\text{IJCNN} & 2 & 125344 & 31337 & 22 & 0.0954 \\
		\text{Porto-Seguro} & 2 & 476169 & 119043 & 57 & 0.0368 \\
		\text{Santander} & 2 & 160000 & 40000 & 200 & 0.1004 \\
		\text{Skin} & 2 & 196045 & 49012 & 3 & 0.7919 \\
		\text{Susy} & 2 & 4000000 & 1000000 & 18 & 0.4577 \\														
		\midrule
	\end{tabular}
\end{table}

Furthermore, to improve the convergence stability we clipped the gradients by a norm of $10^{-5}$, which is necessary due to the steep curvature of the surrogate losses $\hat \ell$, caused by approximating the step-wise true loss surfaces $\ell$. Data batches were drawn in a stratified random fashion (50\% positive and 50\% negative) with a mini-batch size $N=100$. The update steps were chosen as $K_\alpha=3$ and $K_\beta=10$ and Algorithm~\ref{alg:learning} was run for $300000$ iterations. We conducted experiments with 7 binary classification measures, namely AUC, Equal Error Rate (EER), Average Precision (AP), Miss-classification rate (MCR), F1, Mathew Correlation Coefficient (MCC), Jaccard Coefficient (JAC). All the measures are converted to a loss by a $1-x$ conversion, except for EER and MCR. For the losses that demand converting the predictions $\hat y$ into binary values (e.g. MCR, F1, MCC), we used a thresholding $\hat y \ge \gamma$, where $\gamma$ was optimized for each dataset and loss through cross-validation. 

\subsection{Ablation of Surrogate Models}

This section addresses whether the surrogate should be trained in a universal manner (Section~\ref{sec:universalsurrogate}), or whether they should be optimized on a per-dataset basis following the bilevel programming of Section~\ref{sec:bileveloptim}. For this reason, we designed three different modalities for surrogate learning:

\begin{itemize}
	\item \textbf{Universal Surrogate (SL-U):} Learn the surrogate $\hat \ell$ using Equation~\ref{eq:LUniversal} with hyperparameters $p=0.5$ (due to stratified sampling), $\mu=0, \sigma=1$, and then train only the prediction model $f$ by minimizing $\hat \ell$ through Equation~\ref{eq:LHat}. 
	\item \textbf{Learned from Scratch (SL-S):} Initialize the surrogate network $\hat \ell$ from scratch (randomly), then optimize both Equations~\ref{eq:LHat}-\ref{eq:LApprox} using Algorithm~\ref{alg:learning}. 
	\item \textbf{Refined Surrogate (SL-R):} Initialize the surrogate with the universal solution of Equation~\ref{eq:LUniversal}, and then refine the surrogate by optimizing both Equations~\ref{eq:LHat}-\ref{eq:LApprox} using Algorithm~\ref{alg:learning}.
\end{itemize}

\begin{figure}[!h]
	\centering
	\includegraphics[width=1.0\linewidth, trim={0.7cm 4.5cm 0.4cm 0.4cm}]{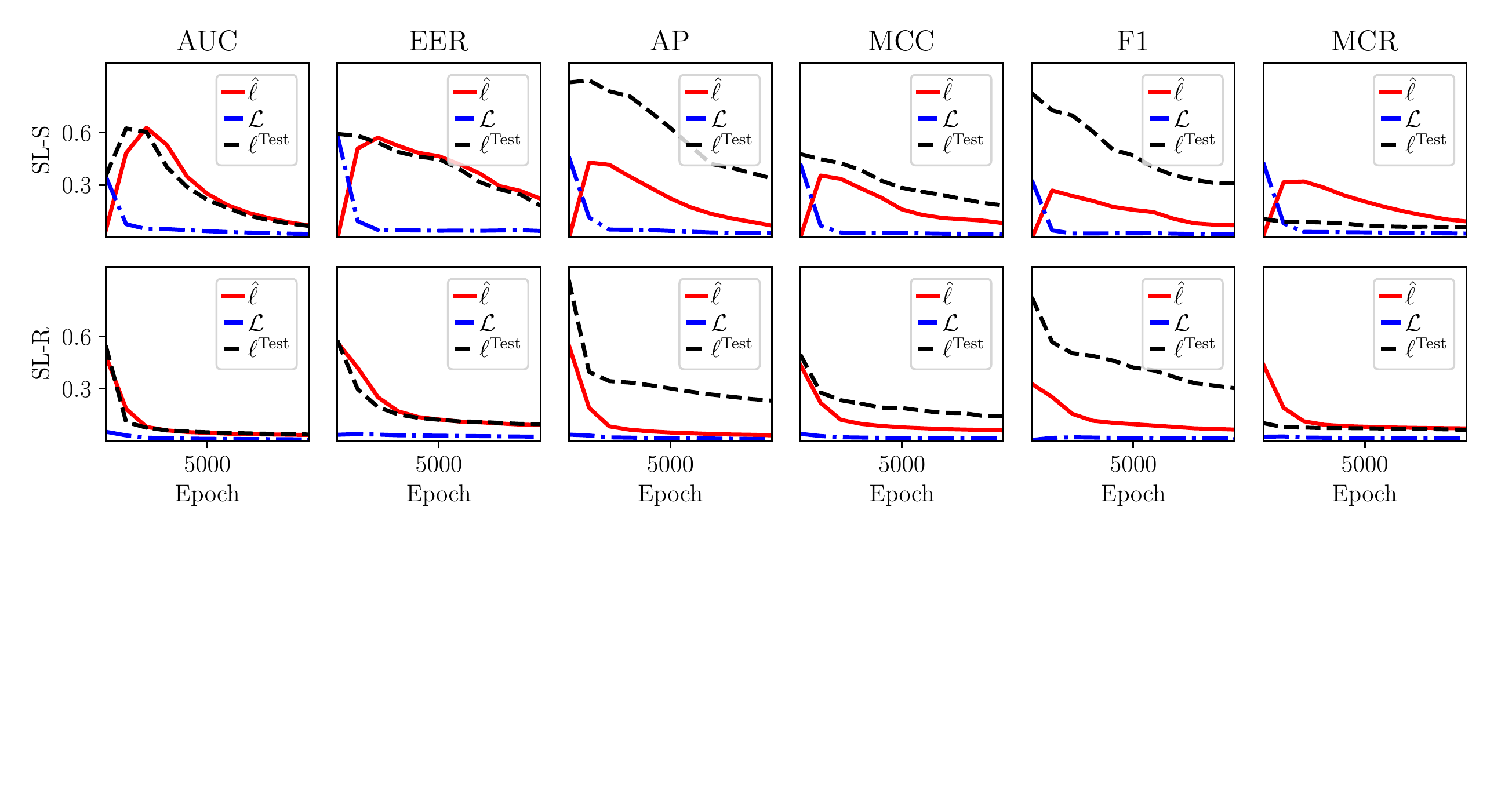}
	\caption{Illustrating the convergence for different loss functions using the IJCNN dataset, using two types of surrogates, SL-S: randomly from scratch, SL-R: refined from the universal surrogate. $\hat \ell$ and $\mathcal{L}$ represent the surrogate optimization performance on the training set, while $\ell^{\text{Test}}$ the true loss on the tesing set.} 
	\label{fig:convergence}
\end{figure}

Table~\ref{tab:results} shows the results of the ablation study with the 3 variations of surrogate learning on 7 losses and 9 datasets. We notice that the universal surrogates (SL-U) are overall sub-optimal with respect to the ones trained in a per-dataset manner (SL-R, SL-S), except for the AUC loss. In addition, the results indicate that there is not a major difference in terms of accuracy between SL-R and SL-U. Nevertheless, the refined surrogate improves the convergence in the early stages of the optimization procedure, as Figure~\ref{fig:convergence} shows. Therefore, we recommend the refined surrogate to a practitioner.

\begin{table}[t]
	\caption{Comparing universal surrogates (SL-U) against dataset-specific surrogates (SL-S and SL-R), which are either initialized randomly (SL-S) or with the universal surrogate (SL-R). The lowest loss values are highlighted.}
	\label{tab:results}
	\small
	\centering
	\begin{tabular}{lcccccccc}
		\toprule 
		\multirow{2}{*}{\textbf{Dataset}} & \multirow{2}{*}{\textbf{Model}} & \multicolumn{7}{c}{\textbf{Loss Measures}} \\ \cmidrule{3-9}
		 & & \textbf{AUC}    & \textbf{EER}    & \textbf{AP}     & \textbf{MCR}    & \textbf{F1}     & \textbf{MCC}    & \textbf{JAC}    \\ \midrule
		\multirow{3}{*}{A9A} & SL-U & \cellcolor{blue!15}0.0968 & 0.1897 & 0.2629 & 0.2165 & 0.4557 & 0.2164 & 0.1819 \\
		& \cellcolor{blue!15}SL-R & 0.0983 & 0.1807 & \cellcolor{blue!15}0.2584 & \cellcolor{blue!15}0.1502 & 0.3134 & 0.2085 & \cellcolor{blue!15}0.1512 \\
		& \cellcolor{blue!15}SL-S & 0.0969 & \cellcolor{blue!15}0.1782 & 0.2585 & 0.1508 & \cellcolor{blue!15}0.3115 & \cellcolor{blue!15}0.2069 & 0.1529 \\ \midrule
		\multirow{3}{*}{CC-Fraud} & SL-U & 0.0245 & 0.1512 & 0.2530 & 0.0093 & 0.7774 & 0.3975 & 0.0099 \\
		& \cellcolor{blue!15}SL-R & 0.0284 & 0.0914 & \cellcolor{blue!15}0.1650 & \cellcolor{blue!15}0.0088 & \cellcolor{blue!15}0.7693 & \cellcolor{blue!15}0.3293 & \cellcolor{blue!15}0.0088 \\
		& SL-S & \cellcolor{blue!15}0.0209 & \cellcolor{blue!15}0.0814 & 0.1902 & \cellcolor{blue!15}0.0088 & 0.9970 & \cellcolor{blue!15}0.3293 & 0.0089 \\ \midrule
		\multirow{3}{*}{Cod-RNA} & SL-U & 0.0107 & 0.0513 & 0.0293 & 0.0510 & 0.1430 & 0.0514 & 0.1272 \\
		& SL-R & 0.0110 & \cellcolor{blue!15}0.0422 & \cellcolor{blue!15}0.0273 & 0.0430 & 0.0632 & 0.0476 & 0.0426 \\
		& \cellcolor{blue!15}SL-S & \cellcolor{blue!15}0.0101 & 0.0428 & 0.0275 & \cellcolor{blue!15}0.0418 & \cellcolor{blue!15}0.0619 & \cellcolor{blue!15}0.0474 & \cellcolor{blue!15}0.0422 \\ \midrule
		\multirow{3}{*}{Covtype} & SL-U & \cellcolor{blue!15} 0.1450 & 0.3019 & 0.1730 & 0.3466 & 0.2699 & 0.2349 & 0.2886 \\
		& \cellcolor{blue!15}SL-R & 0.1463 & \cellcolor{blue!15}0.2220 & 0.1663 & \cellcolor{blue!15}0.2198 & \cellcolor{blue!15}0.2102 & \cellcolor{blue!15}0.2165 & \cellcolor{blue!15}0.2192 \\
		& SL-S & 0.1468 & 0.2233 & \cellcolor{blue!15}0.1635 & 0.2207 & 0.2109 & 0.2203 & 0.2219 \\ \midrule
		\multirow{3}{*}{IJCNN} & SL-U & 0.0048 & 0.0378 & 0.0537 & 0.1045 & 0.5486 & 0.1923 & 0.0609 \\
		& SL-R & 0.0030 & 0.0268 & 0.0224 & 0.0161 & 0.0806 & \cellcolor{blue!15}0.0422 & \cellcolor{blue!15}0.0152 \\
		& \cellcolor{blue!15}SL-S & \cellcolor{blue!15}0.0028 & \cellcolor{blue!15}0.0260 & \cellcolor{blue!15}0.0219 & \cellcolor{blue!15}0.0153 & \cellcolor{blue!15}0.0777 & 0.0427 & 0.0161 \\ \midrule
		\multirow{3}{*}{Porto-Seguro} & SL-U & 0.3745 & 0.4348 & 0.9376 & 0.0461 & 0.8990 & 0.4901 & 0.0459 \\
		& \cellcolor{blue!15}SL-R & \cellcolor{blue!15}0.3737 & \cellcolor{blue!15}0.4110 & \cellcolor{blue!15}0.9361 & 0.0449 & \cellcolor{blue!15}0.8867 & \cellcolor{blue!15} 0.4606 & 0.0448 \\
		& SL-S & 0.3744 & 0.4127 & 0.9367 & \cellcolor{blue!15} 0.0446 & 0.8939 & 0.4614 & \cellcolor{blue!15} 0.0446 \\ \midrule
		\multirow{3}{*}{Santander} & SL-U & \cellcolor{blue!15}0.1453 & 0.2349 & 0.5027 & 0.1082 & 0.5800 & 0.3012 & 0.0951 \\
		& SL-R & 0.1465 & \cellcolor{blue!15}0.2300 & 0.4996 & \cellcolor{blue!15}0.0864 & 0.5166 & 0.2863 & 0.0863 \\ 
		& \cellcolor{blue!15}SL-S & 0.1470 & 0.2314 & \cellcolor{blue!15}0.4945 & 0.0871 & \cellcolor{blue!15}0.5083 & \cellcolor{blue!15}0.2857 & \cellcolor{blue!15}0.0861 \\ \midrule
		\multirow{3}{*}{Skin} & SL-U & \cellcolor{blue!15}0.0001 & 0.0648 & \cellcolor{blue!15}0.0001 & 0.0075 & 0.0036 & 0.0836 & 0.0889 \\
		& SL-R & 0.0006 & \cellcolor{blue!15}0.0044 & 0.0003 & 0.0073 & 0.0085 & \cellcolor{blue!15}0.0031 & 0.0115 \\
		& \cellcolor{blue!15}SL-S & 0.0009 & 0.0143 & \cellcolor{blue!15}0.0001 & \cellcolor{blue!15}0.0047 & \cellcolor{blue!15}0.0017 & 0.0035 & \cellcolor{blue!15}0.0031 \\ \midrule
		\multirow{3}{*}{Susy} & SL-U & \cellcolor{blue!15}0.1316 & 0.2269 & 0.1312 & 0.4596 & 0.2507 & 0.2133 & 0.2461 \\
		& \cellcolor{blue!15}SL-R & 0.1334 & 0.2156 & \cellcolor{blue!15}0.1288 & 0.2046 & \cellcolor{blue!15}0.2289 & \cellcolor{blue!15}0.2050 & \cellcolor{blue!15}0.2031 \\
		& SL-S & 0.1324 & \cellcolor{blue!15}0.2140 & 0.1291 & \cellcolor{blue!15}0.2043 & 0.2294 & 0.2062 & 0.2050 \\ \bottomrule
		\multirow{3}{*}{\textbf{Ranks}}  & SL-U & \cellcolor{green!15}\textbf{1.78}   & 2.89   & 2.67   & 2.78   & 2.67   & 2.89   & 2.78   \\
		& SL-R & 2.22   & 1.67   & \cellcolor{green!15}\textbf{1.67}   & 1.78   & 1.78   & \cellcolor{green!15}\textbf{1.44}   & \cellcolor{green!15}\textbf{1.56}   \\
		& \cellcolor{green!15}SL-S & 1.89   & \cellcolor{green!15}\textbf{1.44}   & \cellcolor{green!15}\textbf{1.67}   & \cellcolor{green!15}\textbf{1.22}   & \cellcolor{green!15}\textbf{1.67}   & 1.67   & 1.67 \\ \bottomrule
	\end{tabular}
\end{table}

\subsection{Comparison with the State-of-the-art}
\label{sec:sotacomp}

In order to demonstrate the usefulness of learning surrogates, we will compare our method against the state-of-the art baselines of four popular loss functions: MCR, AUC, F1 and JAC. Concretely, cross-entropy is the relaxation of MCR, while Pairwise Ranking~\cite{gao2015consistency,NIPS2009_3708} and Global Objective~\cite{pmlr-v54-eban17a} are the relaxations of AUC. Furthermore, the Lovasz Soft-Max is the relaxation of JAC~\cite{berman2018lovasz,pmlr-v37-yub15}, and the cost-sensitive reduction~\cite{NIPS2014_5508} serves as the surrogate of F1. All the aforementioned baselines, except the cost-sensitive reduction, have no further hyper-parameters and were implemented based on the authors' codes. For a \textit{ceteris paribus} comparison, we used the same capacity prediction model for the baselines, the same batch size, i.e. the same protocol as in Section~\ref{sec:protocol}. The hyper-parameter of the cost-sensitive reduction, namely the positive weight coefficient was tuned among $\{ 0.3, 0.9, 2.7, 8.1, 24.3, 72.9 \}$ on a separate validation set. To ensure that the baselines converged, we trained them for $1M$ iterations with a learning rate of $10^{-4}$. Table~\ref{tab:comparisonsota} presents the results over the 9 datasets, where the refined surrogate learning SL-R is compared to the 5 state-of-the-art relaxation methods for 4 losses. The evidence suggests that surrogate learning yields more accurate prediction models than the state-of-the-art.

\begin{table}[ht!]
	\caption{Surrogate learning SL-R vs state-of-the-art, \textbf{MCR}: CE (Cross-Entropy); \textbf{AUC}: PR (Pairwise Ranking~\cite{gao2015consistency,NIPS2009_3708}), GO (Global Objectives~\cite{pmlr-v54-eban17a}); \textbf{JAC}: LO (Lovasz Soft-Max for Jaccard~\cite{berman2018lovasz,pmlr-v37-yub15}); \textbf{F1}: CS (Cost-sensitive F1 reduction~\cite{NIPS2014_5508}). Lowest values in bold. } 
	\centering
	\small
	\begin{tabular}{cccccccccc}
		\toprule 
		\multirow{2}{*}{\textbf{Data}} & \multicolumn{2}{c}{\textbf{MCR}} & \multicolumn{3}{c}{\textbf{AUC}} & \multicolumn{2}{c}{\textbf{JAC}} & \multicolumn{2}{c}{\textbf{F1}} \\ \cmidrule(lr){2-3} \cmidrule(lr){4-6} \cmidrule(lr){7-8} \cmidrule(lr){9-10}
		 & \textbf{CE}    & \textbf{SL-R}    & \textbf{PR}~\cite{gao2015consistency} & \textbf{GO}~\cite{pmlr-v54-eban17a}    & \textbf{SL-R} & \textbf{LO}~\cite{berman2018lovasz} & \textbf{SL-R} & \textbf{CS}~\cite{NIPS2014_5508} & \textbf{SL-R} \\ \cmidrule(lr){1-1} \cmidrule(lr){2-2} \cmidrule(lr){3-3} \cmidrule(lr){4-4} \cmidrule(lr){5-5} \cmidrule(lr){6-6} \cmidrule(lr){7-7} \cmidrule(lr){8-8} \cmidrule(lr){9-9} \cmidrule(lr){10-10} 
		A9A & 0.1520 & \cellcolor{blue!15}0.1502 & 0.1019 & 0.1028 & \cellcolor{blue!15}0.0983 & 0.1539 & \cellcolor{blue!15}0.1512 & 0.3177 & \cellcolor{blue!15}0.3134 \\ 
		CCF & \cellcolor{blue!15}0.0088 & \cellcolor{blue!15}0.0088 & 0.0437 & 0.0369 & \cellcolor{blue!15}0.0284 & \cellcolor{blue!15}0.0088 & \cellcolor{blue!15}0.0088 & \cellcolor{blue!15}0.7652 & 0.7693  \\
		COD & 0.0462 & \cellcolor{blue!15}0.0430 & 0.0122 & 0.0129 & \cellcolor{blue!15}0.0110 & 0.0438 & \cellcolor{blue!15}0.0426 & 0.0652 & \cellcolor{blue!15}0.0632 \\
		COV & \cellcolor{blue!15}0.2149 & 0.2198 & 0.1786 & 0.1504 & \cellcolor{blue!15}0.1463 & 0.2594 & \cellcolor{blue!15}0.2192 & 0.2305 & \cellcolor{blue!15}0.2102 \\
		IJC & 0.0364 & \cellcolor{blue!15}0.0161 & 0.0168 & 0.0258 & \cellcolor{blue!15}0.0030 & 0.0322 & \cellcolor{blue!15}0.0152 & 0.1959 & \cellcolor{blue!15}0.0806 \\
		POR & \cellcolor{blue!15}0.0446 & 0.0449 & 0.3814 & 0.3815  & \cellcolor{blue!15}0.3737 & \cellcolor{blue!15}0.0445 & 0.0448 & \cellcolor{blue!15}0.8851 & 0.8867 \\
		SAN & \cellcolor{blue!15}0.0842 & 0.0864 & \cellcolor{blue!15}0.1406 & 0.1427 & 0.1465 & \cellcolor{blue!15}0.0850 & 0.0863 & \cellcolor{blue!15}0.4990 & 0.5166 \\
		SKI & 0.0482 & \cellcolor{blue!15}0.0073 & 0.0364 & 0.0473 & \cellcolor{blue!15}0.0006 & 0.0432 & \cellcolor{blue!15}0.0115 & 0.0278 & \cellcolor{blue!15}0.0085 \\
		SUS & 0.2146 & \cellcolor{blue!15}0.2046 & 0.1524 & 0.1508 & 0.1334 & \cellcolor{blue!15}0.2022 & 0.2031  & 0.2420 & \cellcolor{blue!15}0.2289 \\ \midrule
		\textbf{Wins} & 3.5 & \bf \cellcolor{green!15}5.5 & 1.0 & 0.0 & \bf \cellcolor{green!15}8.0 & 3.5 & \bf \cellcolor{green!15}5.5 & 3.0 & \bf \cellcolor{green!15}6.0 \\ \bottomrule 
	\end{tabular} 
	\label{tab:comparisonsota}
\end{table}

\subsection{Runtime Complexity}

Denoting the capacities as $\alpha \in \R^{Q_\alpha}, \beta \in \R^{Q_\beta}$, the runtime complexity of Algorithm~\ref{alg:learning} is $\mathcal{O}\left(T \times \left( K_\alpha \times \left(Q_\alpha + Q_\beta\right) + K_\beta \times Q_\beta \right) \right)$, while for comparison that of gradient descent for minimizing the cross-entropy (CE) is $\mathcal{O}\left(T \times K_\alpha \times Q_\alpha \right)$. The additive complexity comes from $\frac{\partial \hat \ell}{\partial \hat y}\frac{\partial \hat y}{\partial \alpha}$ where $\mathcal{O}\left(\frac{\partial \hat \ell}{\partial \hat y}\right)$ is $\mathcal{O}\left(Q_\beta\right)$ in the case of surrogate learning. For the largest dataset Susy it took SL circa 1 day and 4 hours of training time on a single Nvidia 1080 Ti GPU. The baselines were trained on CPU machines due to limited GPU resources, therefore the run-times are not directly comparable (e.g. CE took 2 days and 21 hours for Susy using 20 CPUs). Overall, we noticed that learning surrogates is feasible in practice.

\section{Conclusion}

The optimization of losses is a major challenge for the machine learning community. Unfortunately, most classification loss functions are only piece-wise continuous, non-differentiable and non-decomposable. So far, researchers have addressed this bottleneck by designing (hand-crafting) smooth approximative surrogate functions to those losses. In contrast to the existing work, we propose a new paradigm to optimizing loss functions, by defining the loss itself as a parametric model that is jointly optimized with a prediction model, in a way that the smooth surrogate loss matches the non-differentiable true loss. The task is formalized as a bilevel programming objective and an alternating optimization algorithm is applied to learn the surrogates. The empirical results on multiple real-life datasets indicate that learning surrogates is more accurate than hand-crafted explicit relaxations in diverse popular loss functions, such as AUC, F1, or Jaccard Index. 

\vfill
\newpage
\bibliographystyle{plainnat}
\bibliography{surrogaterefs}

\end{document}